\documentclass{article} 

\usepackage{gen2_iclr2026,times}
\usepackage{booktabs}
\usepackage{multirow}
\usepackage{graphicx}
\usepackage{algorithm}
\usepackage{algorithmic}
\usepackage{amsmath}
\usepackage{amssymb}

\usepackage{amsmath,amsfonts,bm}

\def\eqref#1{equation~\ref{#1}}

\def\1{\bm{1}}

\DeclareMathAlphabet{\mathsfit}{\encodingdefault}{\sfdefault}{m}{sl}
\SetMathAlphabet{\mathsfit}{bold}{\encodingdefault}{\sfdefault}{bx}{n}

\newsavebox\CBox
\def\textBF#1{\sbox\CBox{#1}\resizebox{\wd\CBox}{\ht\CBox}{\textbf{#1}}}

\usepackage{url}
\usepackage[hidelinks]{hyperref}

\title{Protein Counterfactuals via Diffusion-Guided Latent Optimization}

\author{
Weronika Kłos\textsuperscript{1,2} \quad
Sidney Bender\textsuperscript{1,2} \quad
Lukas Kades\textsuperscript{3} \\[0.5em]
\textsuperscript{1}Machine Learning Group, Technische Universität Berlin, Berlin, Germany \\
\textsuperscript{2}Berlin Institute for the Foundations of Learning and Data (BIFOLD) \\
\textsuperscript{3}BASF Digital Solutions GmbH, Ludwigshafen am Rhein, Germany \\[0.3em]
\texttt{\{w.klos,s.bender\}@tu-berlin.de} \quad
\texttt{lukas.kades@basf.com}
}
\iclrfinalcopy
\begin{document}

\maketitle

\begin{abstract}
Deep learning models can predict protein properties with unprecedented accuracy but rarely offer mechanistic insight or actionable guidance for engineering improved variants. When a model flags an antibody as unstable, the protein engineer is left without recourse: which mutations would rescue stability while preserving function?
We introduce \textbf{Manifold-Constrained Counterfactual Optimization for Proteins (MCCOP)}, a framework that computes minimal, biologically plausible sequence edits that flip a model's prediction to a desired target state.
MCCOP operates in a continuous joint sequence--structure latent space and employs a pretrained diffusion model as a manifold prior, balancing three objectives: validity (achieving the target property), proximity (minimizing mutations), and plausibility (producing foldable proteins).
We evaluate MCCOP on three protein engineering tasks -- GFP fluorescence rescue, thermodynamic stability enhancement, and E3 ligase activity recovery -- and show that it generates sparser, more plausible counterfactuals than both discrete and continuous baselines.
The recovered mutations align with known biophysical mechanisms, including chromophore packing and hydrophobic core consolidation, establishing MCCOP as a tool for both model interpretation and hypothesis-driven protein design. Our code is publicly available at \href{https://github.com/weroks/mccop}{\texttt{github.com/weroks/mccop}}.
\end{abstract}

\section{Introduction}
Deep learning has transformed computational protein science. Structure prediction models achieve near-experimental accuracy~\citep{jumper2021highly, abramson2024accurate}, protein language models capture evolutionary grammar~\citep{lin2023evolutionary, esm2024esm}, and generative frameworks design novel folds from scratch~\citep{watson2023novo, ingraham2023illuminating}. Yet these models remain oracles rather than guides: when a predictor flags a candidate as ``aggregation-prone'', the engineer receives no indication of which mutations would resolve the problem.

This paper addresses the need for algorithmic recourse: given a protein $P$ predicted to lack a desired property $y_{\text{target}}$, what is the minimal modification such that the prediction changes? This maps directly to \emph{counterfactual explanations}~\citep{wachter2017counterfactual}. Applied to a model of uncertain quality, counterfactuals expose reliance on spurious correlations; applied to a robust model, they generate testable hypotheses for wet-lab validation.

Translating counterfactual methods to proteins introduces two fundamental challenges. First, the \emph{manifold constraint}: Unlike images, proteins are governed by strict epistatic constraints -- a single core mutation can abolish folding while a compensatory mutation restores it. Naive gradient optimization produces adversarial or invalid examples that satisfy the predictor but correspond to unfoldable proteins. Second, discreteness and geometry: Proteins are \emph{discrete sequences} whose function emerges from \emph{continuous 3D geometry}. Gradient-based methods require continuous relaxation, while naively treating them as sequences ignores spatial relationships: a mutation can be compensatory for another one only if the residues are proximal in 3D space, a property not directly apparent from the 1D sequence.

We address both challenges with \textbf{MCCOP}, a gradient-based framework operating in a continuous joint sequence--structure embedding space that uses a pretrained diffusion model as a manifold prior. Our contributions are:
\begin{enumerate}
    \item \textbf{Framework.} MCCOP combines predictor-guided gradient descent with diffusion-based manifold projection and gradient-sensitivity masking to produce sparse, valid, and plausible protein counterfactuals, without task-specific retraining of the generative model.
    \item \textbf{Quantitative evaluation.} On three benchmarks, MCCOP achieves near-perfect success rates with 3--5$\times$ fewer mutations than discrete baselines and near-zero adversarial rates.
    \item \textbf{Mechanistic interpretability.} MCCOP rediscovers known functional motifs and in several cases exactly recovers ground-truth counterfactual sequences from held-out test data.
\end{enumerate}

\begin{figure}[h]
\begin{center}
\includegraphics[width=\linewidth]{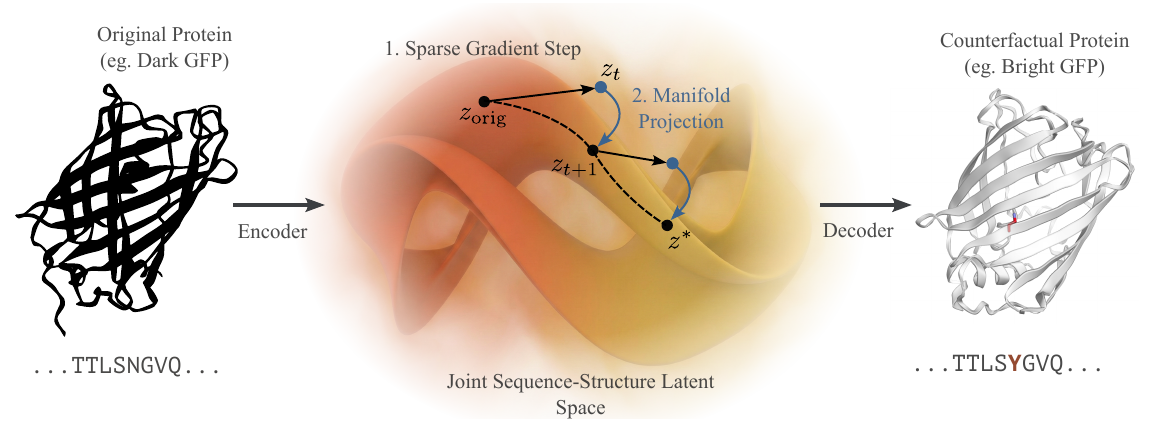}
\end{center}
\caption{Overview of MCCOP. A non-fluorescent GFP variant is mapped to a continuous joint sequence–structure latent space via a pretrained autoencoder. After smoothing the classification boundary, the counterfactual embedding is optimized by alternating between (1) a sparse gradient step maximizing target class probability and (2) manifold projection using a pretrained diffusion model (DiMA). The counterfactual shown is a rediscovered sample from the held-out test set (red sticks denote the mutated residue).}
\label{fig:abstract}
\end{figure}

An overview of our approach is depicted in Figure \ref{fig:abstract}.

\section{Related Work}

\paragraph{Protein language models and embeddings.}
Models such as ESM-2~\citep{lin2023evolutionary} and ESM-C~\citep{esm2024esm} learn unsupervised representations from millions of sequences. Recent multimodal embeddings go further: CHEAP~\citep{lu2025tokenized} compresses ESMFold~\citep{lin2023evolutionary} activations into a joint sequence--structure representation whose decoder maps back to both amino acid sequences and atomistic coordinates. This bidirectional mapping is central to our approach.

\paragraph{Diffusion and generative protein design.}
EvoDiff~\citep{alamdari2023protein} and DiMA~\citep{meshchaninov2024diffusion} apply diffusion to discrete sequences or continuous embeddings; RFdiffusion~\citep{watson2023novo} and folding diffusion~\citep{wu2024protein} operate in $SE(3)$ space (see~\citet{wang2025toward} for an overview). Most generative methods focus on conditional or unconditional sampling. MCCOP differs by using diffusion not for generation but as a regularizer within an optimization loop -- conceptually an inversion of classifier guidance.

\paragraph{Explainability in the protein domain.}
Prior work relies on attention visualization~\citep{vig2020bertology}, feature attribution~\citep{sibli2025enhancing, dickinson2022positional}, gradient-based structure perturbation~\citep{tan2023explainablefold}, or sparse autoencoders applied to pLMs~\citep{gujral2025sparse}. Unlike passive attribution, MCCOP provides \emph{active recourse}: not just why a protein is predicted to fail, but how to rescue it.

\paragraph{Counterfactual explanations.}
Counterfactuals, formalized for ML by~\citet{wachter2017counterfactual}, seek minimal input modifications that change a model's output -- a concept not to be confused with causal counterfactual inference in the structural causal model (SCM) sense~\citep{pearl2009causality}. Methods for tabular data~\citep{mothilal2020explaining, russell2019efficient} are well established. For high-dimensional inputs, diffusion-based approaches (DVCE~\citep{augustin2022diffusion}, DIME~\citep{jeanneret2022diffusion}, Diff-ICE~\citep{pegios2025diffusion}, FastDiME~\citep{weng2024fast}, ACE~\citep{jeanneret2023adversarial}) generate on-manifold counterfactuals via guided denoising, with extensions to diverse sets~\citep{bender2025towards, bender2026visual}, graphs~\citep{bechtoldt2025graph, chen2023d4explainer}, and text~\citep{sarkar2024large}. GAN/VAE-based predecessors include DiVE~\citep{rodriguez2021beyond} and Diffeomorphic Counterfactuals~\citep{dombrowski2024diffeomorphic}. To our knowledge, no prior work applies diffusion-guided counterfactual optimization to proteins. The closest biological relatives -- latent fitness optimization~\citep{ngo2024latent, castro2022relso} -- seek global optima rather than minimal edits and train task-specific generative models.

\section{Methods}
\label{sec:methods}
We now describe each component of MCCOP: the latent representation, predictor smoothing, and the counterfactual optimization loop itself.

\subsection{Problem Formulation}
Let $\mathcal{M} \subset \mathbb{R}^{L' \times D}$ denote the manifold of biologically plausible protein embeddings. Given a predictor $f_\theta: \mathcal{M} \to \mathcal{Y}$ and an input embedding $z_0 \in \mathcal{M}$ with prediction $y_0 = f_\theta(z_0)$, we seek:
\begin{equation}
    z^* = \arg\min_{z \in \mathcal{M}} \left[ \mathcal{L}_{\text{task}}(f_\theta(z),\, y_{\text{target}}) + \lambda\, d(z,\, z_0) \right],
    \label{eq:cf-objective}
\end{equation}
where $d$ enforces proximity to $z_0$ and $z \in \mathcal{M}$ ensures plausibility. Without the manifold constraint, optimization yields adversarial examples~\citep{dombrowski2024diffeomorphic}. We enforce it implicitly using the score function of a diffusion model trained on protein embeddings, whose denoising step acts as a projection $\Pi_\mathcal{M}$, interleaved with gradient steps on Eq.~\ref{eq:cf-objective}.

\subsection{Latent Representation}
\label{sec:latent}
We map sequences $S \in \mathcal{A}^L$ to continuous representations $z \in \mathbb{R}^{L' \times D}$ using CHEAP~\citep{lu2025tokenized}. The encoder $\mathcal{E}$ compresses ESMFold activations into embeddings jointly capturing evolutionary and structural information. The decoder $\mathcal{D}$ maps $z$ back to both a sequence $\hat{S} = \mathcal{D}_{\text{seq}}(z)$ and backbone coordinates $\hat{\Omega} = \mathcal{D}_{\text{struct}}(z)$, with near-perfect round-trip reconstruction ($>$99\% residue accuracy). Crucially, $\mathcal{D}$ is a position-wise MLP -- each token $\hat{S}_i$ depends only on $z_i$ -- enabling sequence-level sparsity via row-wise latent masking (\S\ref{sec:sparsity}). Both encoder and decoder are frozen throughout.

\subsection{Predictor Smoothing}
\label{sec:smoothing}
Our framework is model-agnostic: any differentiable predictor on CHEAP embeddings can be used. As a test-bed we train a shallow MLP $f_\theta$ on flattened embeddings (architecture details in Appendix~\ref{app:predictor-details}).

A non-smooth $f_\theta$ produces high-frequency gradients guiding optimization toward adversarial perturbations. Motivated by the observation of~\citet{bender2025towards}, that smooth classifiers yield more reliable counterfactual optimization, we smooth $f_\theta$ via four complementary mechanisms: (1)~\textbf{spectral normalization}~\citep{miyato2018spectral} on all linear layers; (2)~\textbf{Jacobian regularization}~\citep{jakubovitz2018improving}, penalizing $\|\nabla_z f_\theta(z)\|_F^2$; (3)~\textbf{Softplus activations} ($\beta=1$); and (4)~\textbf{embedding-space adversarial augmentation} via FGSM~\citep{goodfellow2014explaining}, where perturbations decoding to the original sequence are added with the original label, teaching invariance to semantically null perturbations. As shown in Table~\ref{tab:smoothing-results}, this reduces gradient norms by up to 4$\times$ while maintaining or improving AUROC.

\subsection{Counterfactual Optimization}
\label{sec:cf-optimization}

Given embedding $z_{\text{orig}}$ with predicted class $y_{\text{orig}}$, we seek $z^*$ such that $f_\theta(z^*) = y_{\text{target}} \neq y_{\text{orig}}$, minimizing decoded mutations while staying on $\mathcal{M}$. Algorithm~\ref{alg:mccop} summarizes the procedure.

\subsubsection{Objective Function}
\label{sec:objective}
At step $t$, we minimize:
\begin{equation}
    \mathcal{L}_{\text{CF}}(z_t) = \underbrace{\log\bigl(1 + \exp(m - \tilde{y} \cdot f_\theta(z_t))\bigr)}_{\mathcal{L}_{\text{margin}}} + \lambda_{\text{dist}} \underbrace{\|z_t - z_{\text{orig}}\|_2^2}_{\mathcal{L}_{\text{prox}}}
    \label{eq:cf-loss}
\end{equation}
where $\tilde{y} \in \{-1, +1\}$ is the signed target label, $m > 0$ is a confidence margin, and $\lambda_{\text{dist}}$ controls the proximity-validity trade-off.

\subsubsection{Gradient-Based Sparsity Masking}
\label{sec:sparsity}
We compute per-position sensitivity $s_i = \|\nabla_{z_i} \mathcal{L}_{\text{CF}}\|_2$ and construct a binary mask selecting the top-$k$ positions:
\begin{equation}
    M_i = \mathbf{1}\bigl[s_i \geq s_{(k)}\bigr].
\end{equation}
Gradients are applied only at masked positions; non-masked positions are hard-reset to $z_{\text{orig}}$. Because $\mathcal{D}$ is position-wise, row-wise masking in latent space directly enforces sequence-space sparsity. The mask can alternatively be user-defined for constrained editing (e.g., fixing catalytic residues).

\subsubsection{Manifold Projection}
\label{sec:projection}
We regularize the trajectory using DiMA~\citep{meshchaninov2024diffusion} as an implicit manifold prior. At each step, we partially diffuse to noise level $t_{\text{diff}}$, denoise to obtain $\Pi_\phi(z'_t)$, and blend:
\begin{equation}
    z_{t+1} = (1 - \alpha)\, z'_t + \alpha\, \Pi_\phi(z'_t),
    \label{eq:projection}
\end{equation}
where $\alpha \in [0, 1]$ controls projection strength ($\alpha = 0$: unconstrained; $\alpha = 1$: full projection, which destabilizes optimization). We use $\alpha = 0.3$ in practice (ablation in Appendix~\ref{app:hyperparameters}).

\begin{algorithm}[t]
\caption{MCCOP: Manifold-Constrained Counterfactual Optimization for Proteins}
\label{alg:mccop}
\begin{algorithmic}[1]
\REQUIRE Embedding $z_{\text{orig}}$, predictor $f_\theta$, diffusion projector $\Pi_\phi$, target label $\tilde{y}$, sparsity $k$, projection strength $\alpha$, margin $m$, learning rate $\eta$, max steps $T_{\max}$, confidence threshold $\tau$
\STATE $z_0 \leftarrow z_{\text{orig}}$
\FOR{$t = 0, 1, \ldots, T_{\max} - 1$}
    \STATE Compute $\mathcal{L}_{\text{CF}}(z_t)$ via Eq.~\ref{eq:cf-loss}
    \STATE Compute per-position sensitivity: $s_i = \|\nabla_{z_i} \mathcal{L}_{\text{CF}}\|_2$
    \STATE Construct top-$k$ mask: $M_i = \mathbf{1}[s_i \geq s_{(k)}]$
    \STATE Gradient step: $z'_t = z_t - \eta \cdot (M \odot \nabla_z \mathcal{L}_{\text{CF}})$
    \STATE Hard reset: $z'_t[i] \leftarrow z_{\text{orig}}[i]$ for all $i$ where $M_i = 0$
    \STATE Manifold projection: $z_{t+1} = (1 - \alpha)\, z'_t + \alpha\, \Pi_\phi(z'_t)$
    \IF{$\sigma(\tilde{y} \cdot f_\theta(z_{t+1})) \geq \tau$ \AND $\mathcal{D}_{\text{seq}}(z_{t+1}) \neq S_{\text{orig}}$}
        \STATE \textbf{return} $z_{t+1}$ \COMMENT{Early stopping: valid counterfactual found}
    \ENDIF
\ENDFOR
\STATE \textbf{return} $z_{T_{\max}}$ \COMMENT{Return best attempt}
\end{algorithmic}
\end{algorithm}

\subsection{Experimental Setup}

\subsubsection{Datasets}
We evaluate on three datasets with diverse physical origins (statistics in Appendix~\ref{app:datasets}):
\textbf{(1)~TAPE Fluorescence}~\citep{sarkisyan2016local, rao2019evaluating}: GFP homologs with bimodal fluorescence, binarized into bright/dark classes (optimize dark$\to$bright).
\textbf{(2)~TAPE Stability}~\citep{rocklin2017global}: proteolysis-based stability measurements; we remove the middle 33\% quantile to create stable/unstable classes (optimize unstable$\to$stable).
\textbf{(3)~Ube4b Activity}~\citep{starita2013activity}: $\sim$100k mutations in the U-box domain mapped to auto-ubiquitination activity; middle 33\% removed, active/inactive classes defined by top/bottom quantiles (optimize inactive$\to$active).

\subsubsection{Baselines}
We compare against: (1)~\textbf{Stochastic Hill Climbing}: greedy random single-site mutations; (2)~\textbf{Genetic Algorithm}: population-based evolution with edit-distance-penalized fitness; (3)~\textbf{Gradient Descent}: unconstrained latent optimization without smoothing or manifold projection. Details in Appendix~\ref{app:baslines}.

\subsubsection{Evaluation Metrics}
We assess \textbf{validity and sparsity} via success rate (fraction achieving target class), Hamming distance (number of mutations), and adversarial rate (fraction of successful counterfactuals corresponding to the same sequence). \textbf{Structural plausibility} is evaluated using ESM3-predicted pLDDT confidence and radius of gyration ($R_g$). \textbf{Physicochemical plausibility} is monitored via GRAVY hydrophobicity, instability index, and a binary solubility proxy.

\section{Results}
\label{sec:results}

We evaluate on complete test sets, excluding samples misclassified by the predictor. This results in $n = 2093$, $2209$, and $2600$ samples for the stability, fluorescence, and activity datasets respectively. Results are mean $\pm$ std over three seeds.

\subsection{Predictor Smoothing Improves Robustness Without Sacrificing Accuracy}

\begin{table}[t]
\caption{Predictor AUROC and average $L_2$ gradient norm before and after smoothing (mean $\pm$ std, 3 seeds).}
\label{tab:smoothing-results}
\begin{center}
\begin{tabular}{lcccc}
\multicolumn{1}{c}{Dataset} & \multicolumn{2}{c}{AUROC ($\uparrow$)} & \multicolumn{2}{c}{Avg. $L_2$ norm ($\downarrow$)}\\
 & \it Before & \it After & \it Before & \it After \\ \hline \\
Fluorescence & $0.99 \pm 0.00$ & $0.99 \pm 0.00$ & $2.21 \pm 0.19$ & $1.10 \pm 0.10$ \\
Stability    & $0.94 \pm 0.00$ & $0.98 \pm 0.01$ & $1.38 \pm 0.13$ & $0.36 \pm 0.08$ \\
Activity     & $0.82 \pm 0.00$ & $0.93 \pm 0.01$ & $0.36 \pm 0.06$ & $0.33 \pm 0.04$ \\
\end{tabular}
\end{center}
\end{table}

Table~\ref{tab:smoothing-results} shows that smoothing reduces gradient norms by up to 4$\times$ while maintaining or improving AUROC. The largest gain is on the activity dataset (AUROC: 0.82$\to$0.93), likely because Jacobian regularization and adversarial augmentation reduce overfitting to noisy labels.

\subsection{MCCOP Produces Valid and Sparse Counterfactuals}

\begin{table}[t]
\caption{Success rate, adversarial rate, and edit distance (mean $\pm$ std, 3 seeds). Edit distance computed on successful counterfactuals only (confidence $\geq 0.95$, edit distance $\geq 1$). Discrete methods cannot produce adversarial examples by construction, so no values are bolded in this column. $^\dagger$Gradient Descent achieves 100\% adversarial rate; edit distance is undefined.}
\label{tab:combined-results}
\centering
\begin{tabular}{llccc}
\toprule
\bf Dataset & \bf Method & \bf Success Rate ($\uparrow$) & \bf Adv. Rate ($\downarrow$) & \bf Edit Dist. ($\downarrow$) \\ \midrule
\multirow{4}{*}{Stability} 
 & Genetic Algorithm & $0.55 \pm 0.01$ & $0.00 \pm 0.00$ & $7.76 \pm 0.06$ \\
 & Gradient Descent$^\dagger$ & $1.00 \pm 0.00$ & $1.00 \pm 0.00$ & $-$ \\
 & Stochastic Hill Climb & $0.23 \pm 0.00$ & $0.00 \pm 0.00$ & $9.46 \pm 0.18$ \\
 & \textbf{MCCOP (ours)} & $\textBF{1.00} \pm 0.00$ & $0.03 \pm 0.00$ & $\textBF{2.32} \pm 0.01$ \\
\cmidrule{1-5}
\multirow{4}{*}{Fluorescence}
 & Genetic Algorithm & $\textBF{0.36} \pm 0.30$ & $0.00 \pm 0.00$ & $5.37 \pm 3.09$ \\
 & Gradient Descent$^\dagger$ & $1.00 \pm 0.00$ & $1.00 \pm 0.00$ & $-$ \\
 & Stochastic Hill Climb & $0.13 \pm 0.00$ & $0.00 \pm 0.00$ & $7.79 \pm 0.30$ \\
 & \textbf{MCCOP (ours)} & $0.19 \pm 0.00$ & $0.01 \pm 0.00$ & $\textBF{1.37} \pm 0.01$ \\
\cmidrule{1-5}
\multirow{4}{*}{Activity}
 & Genetic Algorithm & $0.17 \pm 0.15$ & $0.00 \pm 0.00$ & $6.24 \pm 3.67$ \\
 & Gradient Descent$^\dagger$ & $1.00 \pm 0.00$ & $1.00 \pm 0.00$ & $-$ \\
 & Stochastic Hill Climb & $0.03 \pm 0.00$ & $0.00 \pm 0.00$ & $10.91 \pm 0.02$ \\
 & \textbf{MCCOP (ours)} & $\textBF{1.00} \pm 0.00$ & $0.02 \pm 0.02$ & $\textBF{2.46} \pm 0.33$ \\
\bottomrule
\end{tabular}
\end{table}

Table~\ref{tab:combined-results} reveals three key findings. \textbf{(1)~Unconstrained gradient optimization is entirely adversarial:} every counterfactual decodes to the original sequence, confirming exploitable high-frequency artifacts and validating our smoothing and projection pipeline. \textbf{(2)~MCCOP achieves high success with minimal edits:} 100\% success on stability and activity with 2.3--2.5 mutations versus 6.2--10.9 for discrete baselines. MCCOP reaches early stopping after a median of 2--10 steps, while hill climbing exhausts the budget in $>$95\% of cases. \textbf{(3)~Fluorescence is harder:} MCCOP's 19\% success rate reflects the requirement for precise chromophore geometry potentially exceeding our sparsity budget ($k=5$), yet successful counterfactuals are the sparsest (1.4 mutations) with near-zero adversarial rate.

Edit distances for MCCOP and the genetic algorithm are tunable via $k$/$\lambda_{\text{dist}}$ and fitness weighting, respectively; MCCOP's advantage lies in the favorable trade-off between success rate, sparsity, and plausibility.

\subsection{Structural and Physicochemical Plausibility}

\begin{figure}[h]
\begin{center}
\includegraphics[width=0.9\linewidth]{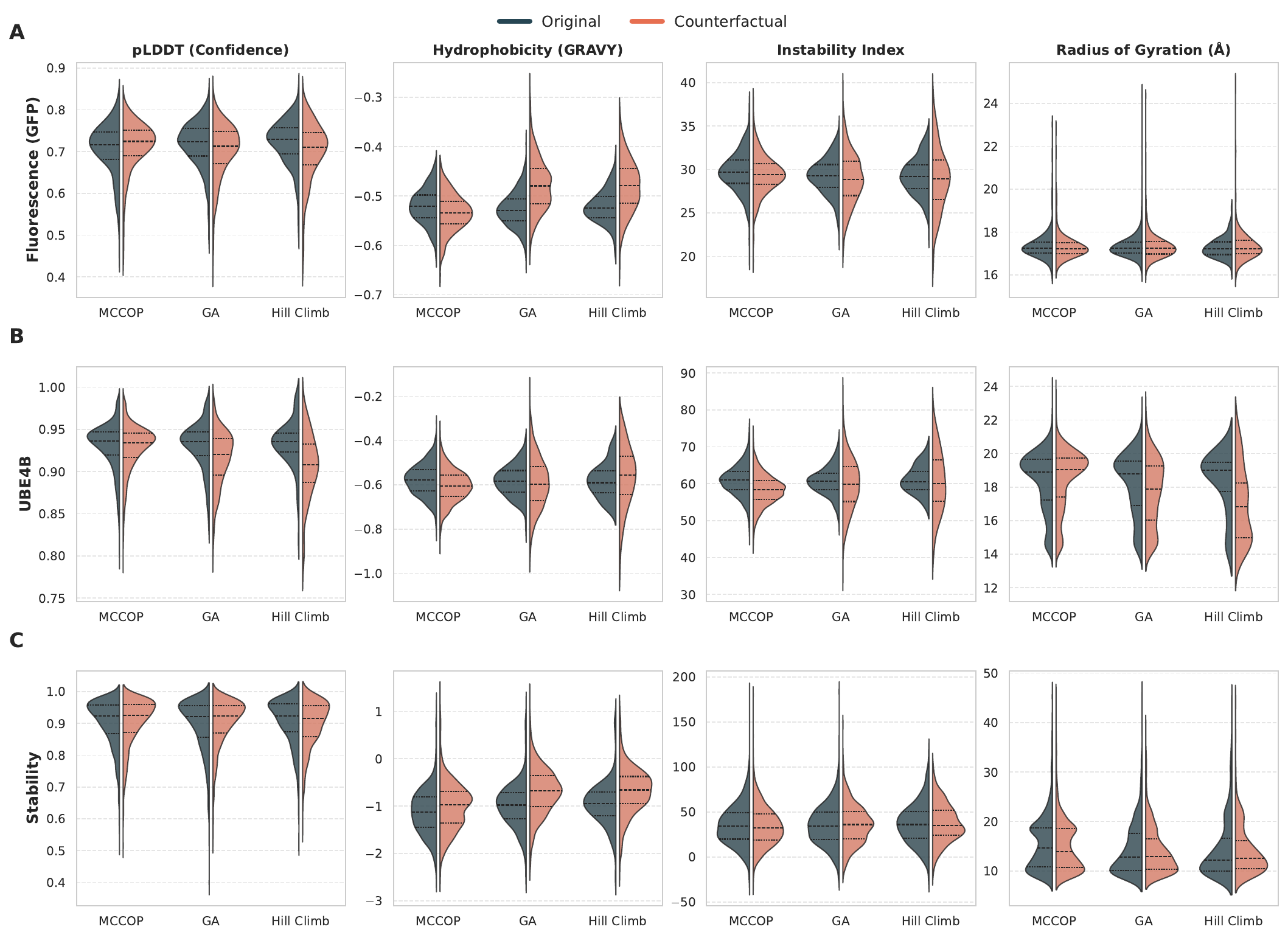}
\end{center}
\caption{Physicochemical plausibility across benchmarks (columns: pLDDT, GRAVY, instability index, $R_g$; rows: fluorescence, activity, stability). MCCOP (orange) closely matches the original distribution (gray); discrete baselines show broader shifts. Statistical comparisons via Kruskal-Wallis/Dunn's tests with Benjamini-Hochberg correction: MCCOP achieves significantly higher pLDDT than both baselines across all tasks (adjusted $p < 0.02$).}
\label{fig:properties}
\end{figure}

Figure~\ref{fig:properties} shows that MCCOP counterfactuals are nearly indistinguishable from the original distribution across all metrics, occasionally shifting toward more favorable values. Discrete baselines introduce broader shifts, especially in hydrophobicity and instability index, as they explore sequence space without structural priors. A controlled comparison at fixed edit distance (Appendix~\ref{app:edit-distance-controlled}) confirms these trends.

\subsection{MCCOP Rediscovers Known Biophysical Mechanisms}

\begin{figure}[h]
\begin{center}
\includegraphics[width=0.9\linewidth]{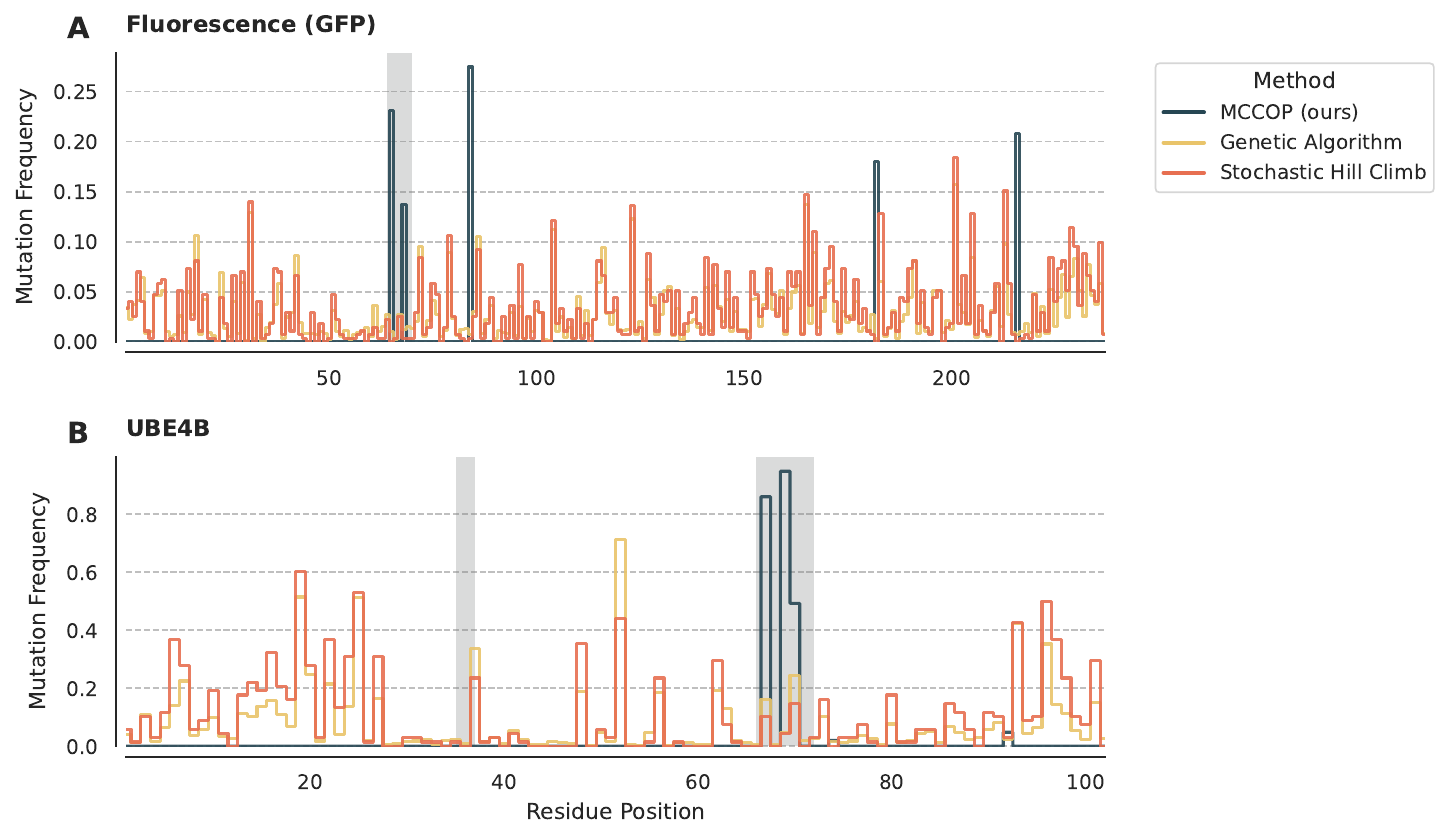}
\end{center}
\caption{Per-residue mutation frequency for fluorescence (\textbf{A}) and Ube4b activity (\textbf{B}). MCCOP (blue) concentrates mutations in functionally relevant regions -- chromophore-proximal residues for GFP, E2-binding interface for Ube4b -- while baselines distribute mutations nearly uniformly. Shaded regions: known functional motifs~\citep{sarkisyan2016local, starita2013activity}.}
\label{fig:mutation_freq}
\end{figure}

\paragraph{GFP fluorescence.}
MCCOP concentrates mutations in the chromophore-proximal region (residues 63--69) and $\beta$-barrel strands forming the chromophore cavity (Figure~\ref{fig:mutation_freq}A), consistent with the requirement for tight packing to suppress non-radiative decay~\citep{sarkisyan2016local}. A small number of distal mutations (e.g., residues 181, 216) may represent novel compensatory interactions or predictor artifacts, requiring experimental follow-up.

\paragraph{Ube4b activity.}
Mutations cluster at the E2-binding interface (residues 66--71; Figure~\ref{fig:mutation_freq}B), through which Ube4b recruits UbcH5c for ubiquitin transfer~\citep{starita2013activity}.

\paragraph{Thermodynamic stability.}
The stability dataset spans diverse topologies~\citep{rocklin2017global}, so no universal residue positions dominate. However, MCCOP frequently targets core-facing residues, suggesting hydrophobic core consolidation as a general stabilization strategy (Appendix~\ref{app:structures}).

\begin{figure}[h]
\begin{center}
\includegraphics[width=0.9\linewidth]{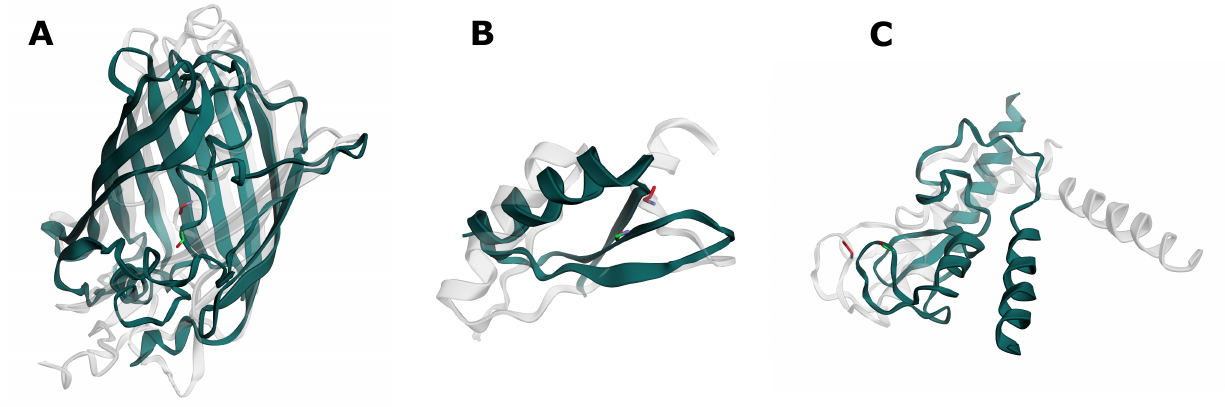}
\end{center}
\caption{Structural alignments between original (gray) and counterfactual (colored) proteins for rediscovered ground-truth examples. (\textbf{A})~GFP: mutations near the chromophore. (\textbf{B})~Stability: core-facing mutations. (\textbf{C})~Ube4b: E2-binding interface mutations. Structures predicted by ESM3.}
\label{fig:rediscovered}
\end{figure}

\paragraph{Recovery of ground-truth counterfactuals.}
MCCOP exactly recovers existing opposite-label sequences in 16 (fluorescence), 18 (activity), and 4 (stability) cases -- several from the held-out test set. Figure~\ref{fig:rediscovered} shows structural alignments confirming that recovered mutations localize to functionally relevant regions.

\section{Discussion}

MCCOP generates sparse, on-manifold counterfactual explanations achieving near-perfect success rates with 1--3 mutations on average (vs.\ 8--11 for discrete baselines) while maintaining structural and physicochemical plausibility. Recovered mutations align with established biophysical mechanisms, suggesting that the underlying predictors have learned meaningful sequence-function relationships.

\paragraph{Explanation versus editing.}
MCCOP's primary goal is model interpretation, not direct engineering. A counterfactual is only as trustworthy as the predictor it explains: if the model has learned spurious correlations, the counterfactual reflects them faithfully -- which is itself diagnostic. When the predictor is robust, MCCOP's outputs become candidates for experimental validation.

\paragraph{From correlation to causation.}
Our framework identifies correlational, not causal, relationships. Establishing true causal links requires interventional experiments, but MCCOP's sparse suggestions (2 mutations vs.\ thousands of directed-evolution variants) are directly amenable to such follow-up.

\paragraph{Limitations.}
(1)~Plausibility evaluation relies on computational proxies (ESM3 pLDDT, $R_g$, physicochemical indices) rather than experimental validation. (2)~The CHEAP encoder--decoder introduces reconstruction error that may produce artifacts for proteins distant from ESMFold's training distribution. (3)~We evaluate only binary tasks; extending to continuous regression targets requires replacing the margin loss with MSE or quantile losses.

\paragraph{On the manifold and smoothness assumptions.}
Two assumptions embedded in our framework deserve scrutiny.
\emph{First}, MCCOP operates in a continuous latent space under the implicit premise that plausible protein sequences concentrate near a low-dimensional manifold.
The same \emph{manifold hypothesis} is routinely invoked in computer vision, where natural images are assumed to populate a thin subspace of pixel space, yet to the best of our knowledge no formal proof of this assertion exists for images or for proteins.
Fefferman et~al.\ develop statistical tests for the hypothesis but do not establish it for any natural data distribution~\citep{fefferman2016testing}; empirically, the evidence is consistent with data concentrating on disconnected clusters or ``blobs'' rather than a single smooth manifold~\citep{bengio2013representation}.
For proteins, the situation is arguably more fraught: functional sequences are constrained by folding, stability, and epistasis, producing a viable sequence space that may be fragmented and topologically complex rather than smoothly connected.
\emph{Second}, MCCOP's Gaussian smoothing of the latent space presupposes that the underlying sequence--function mapping varies smoothly, so that local perturbations yield gradual changes in the predicted phenotype.
However, protein fitness landscapes are known to be rugged: higher-order epistasis creates abrupt fitness transitions even between sequences that differ by a single residue~\citep{weinreich2006darwinian, sarkisyan2016local}, and there is no \emph{a priori} reason to expect the predictor's decision surface, which reflects these landscapes, to be smooth either.
An alternative to smoothing might be signal filtering tuned to a desired frequency, which would suppress high-frequency noise without globally flattening the landscape; we opted for Gaussian smoothing as a pragmatic engineering choice that made the gradient-based optimization tractable, rather than as a theoretically motivated operation.
We flag these points not because they invalidate the results -- MCCOP's strong empirical performance suggests the assumptions are serviceable in practice -- but because they circumscribe the regime in which the method's outputs should be trusted and highlight opportunities for more principled geometric and spectral approaches in future work.

\paragraph{Future directions.}
(1)~\emph{Multi-objective counterfactuals}: jointly optimizing stability and binding affinity by combining predictor gradients.
(2)~\emph{Experimental validation}: synthesizing top-ranked variants for closed-loop validation.
(3)~\emph{Diverse counterfactual sets}~\citep{mothilal2020explaining, bender2026visual}: revealing alternative mutational strategies and enriching fitness landscape understanding.

\subsubsection*{Acknowledgments}

We would like to thank Marvin Sextro for many valuable pieces of advice and proof-reading, as well as Klaus-Robert Müller, Adrian Hill, and Stefan Chmiela for interesting and fruitful discussions. We also would like to thank Dominik Kühne for maintaining our HPC cluster hydra and being always there to help in case of technical difficulties. We used GitHub Copilot for assistance with code development and editing of paper text. All AI-generated content was reviewed, verified, and revised by the authors, who take full responsibility for the final manuscript.

This work was supported by the German Ministry for Education and Research (BMBF) under Grant 01IS18037A, and by BASLEARN -- TU Berlin/BASF Joint Laboratory, co-financed by TU Berlin and BASF SE.

\bibliography{gen2_iclr2026_workshop}
\bibliographystyle{gen2_iclr2026_workshop}

\appendix

\section{Predictor Training and Smoothing Details}
\label{app:predictor-details}

\paragraph{Architecture.}
The property predictor $f_\theta$ is a three-layer MLP with hidden dimensions [512, 256], each followed by spectral normalization~\citep{miyato2018spectral} and Softplus activation ($\beta = 1$). The final layer outputs a single logit. Input embeddings are flattened resulting in an input dimension of sequence length $L$ times embedding dimension $D$ (masking padding tokens) before being passed to the MLP.

\paragraph{Training protocol.}
We train on 80\% of each dataset, reserving 10\% for validation and 10\% for testing, stratified by label. We use Adam with a learning rate of $1 \times 10^{-5}$ and a dropout rate of $0.3$. Early stopping is applied with a patience of 5 epochs based on validation AUROC.

\paragraph{Smoothing mechanisms.}
Further details concerning the smoothing mechanisms include:
\begin{enumerate}
    \item \textbf{Spectral normalization}~\citep{miyato2018spectral}: applied to all linear layers, constraining the Lipschitz constant of each layer to approximately 1.
    \item \textbf{Jacobian regularization}~\citep{jakubovitz2018improving}: we add a penalty term $\lambda_J \| \nabla_z f_\theta(z) \|_F^2$ to the training loss, with $\lambda_J = 10^{-3}$. The Frobenius norm is estimated via a Hutchinson trace estimator with 5 random projections per batch for computational efficiency.
    \item \textbf{Adversarial data augmentation}: for each training sample $(z_i, y_i)$, we generate an adversarial embedding $z_i^{\text{adv}}$ via an FGSM attack ($\epsilon = 0.01$ in embedding space) targeting the opposite class. Adversarial samples that decode to the \emph{same} amino acid sequence as the original (i.e., $\mathcal{D}(z_i^{\text{adv}}) = \mathcal{D}(z_i)$) are added to the training set with the original label $y_i$, teaching the model to be invariant to semantically null perturbations.
\end{enumerate}

\paragraph{Smoothness quantification.}
We report the average $L_2$ gradient norm $\mathbb{E}_{z \sim \mathcal{D}_{\text{test}}} [\| \nabla_z f_\theta(z) \|_2]$ computed over the full test set. Lower values indicate a smoother decision boundary.

\section{Computational Cost Analysis}
\label{app:comp-cost}

\begin{figure}[h]
\begin{center}
\includegraphics[width=0.6\linewidth]{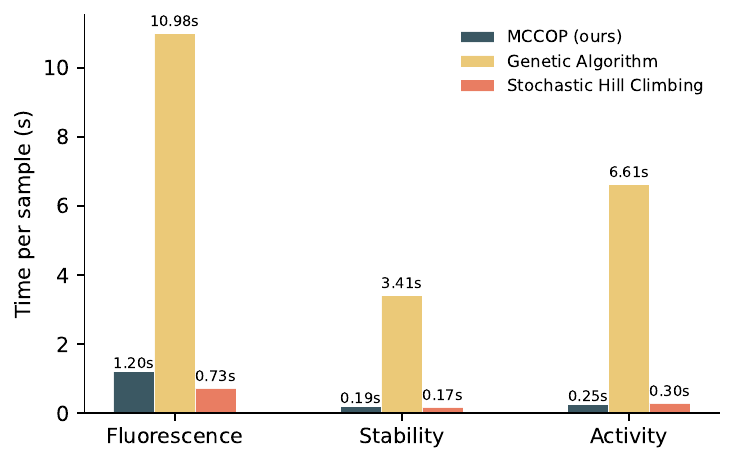}
\end{center}
\caption{Average wall-clock time per sample across the complete test set. The genetic algorithm is the most expensive method. MCCOP and stochastic hill climbing have comparable execution times, though their computational profiles differ: discrete baselines spend $>$94\% of time on re-encoding candidate sequences, while MCCOP spends 99\% on diffusion-based manifold projection.}
\label{fig:comp-cost}
\end{figure}

Figure~\ref{fig:comp-cost} reports the average wall-clock time per sample for MCCOP and the two discrete baselines across all three benchmarks.
The genetic algorithm is the most expensive method by roughly an order of magnitude due to its population-based evaluation.
MCCOP and stochastic hill climbing have comparable per-sample execution times.

An important caveat applies to this comparison.
The discrete baselines operate in sequence space but must evaluate candidates using the same embedding-space predictor; every proposed mutation therefore requires a full re-encoding through the CHEAP encoder (backed by ESMFold), which constitutes 97\% and 94\% of total computation time for hill climbing and the genetic algorithm respectively.
This overhead is not intrinsic to those algorithms but arises from the requirement of a shared evaluation protocol.
MCCOP, by contrast, operates natively in embedding space and avoids this round-trip entirely.
Its dominant cost is the diffusion-based manifold projection, which accounts for 99\% of computation time. For performance-critical applications we would recommend executing the diffusion-based projection step only every $n$ optimization steps.

\section{Controlled Edit Distance Comparison}
\label{app:edit-distance-controlled}

\begin{figure}[h]
\begin{center}
\includegraphics[width=0.9\linewidth]{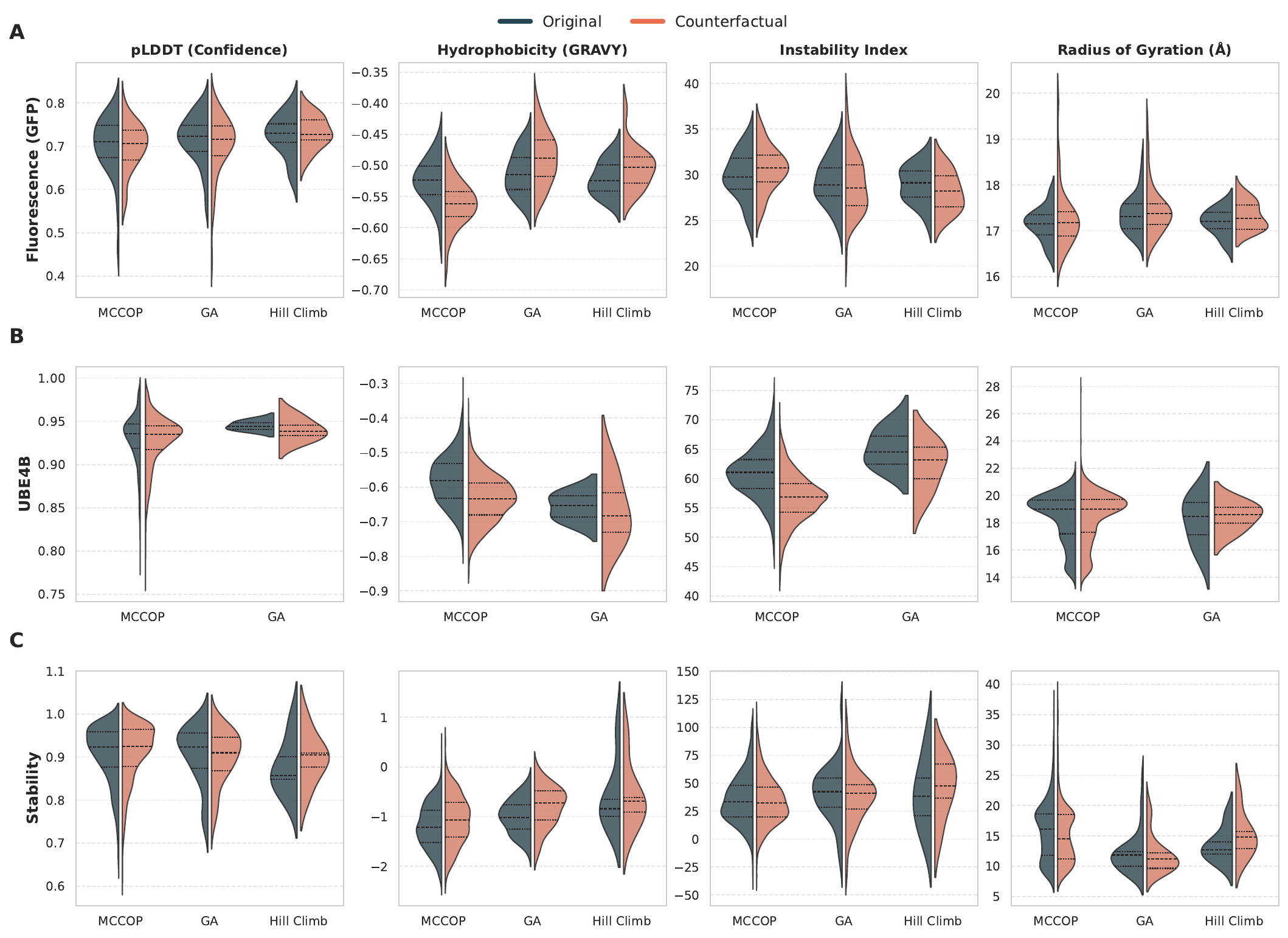}
\end{center}
\caption{Physicochemical property distributions for counterfactuals with exactly 3 mutations. MCCOP closely tracks the original test set distribution, while the genetic algorithm and stochastic hill climbing show greater deviation, particularly for GRAVY and instability index.}
\label{fig:properties-ed3}
\end{figure}

To ensure a fair comparison across methods, we filter all successful counterfactuals to those with exactly three mutations (edit distance = 3), which represents the bin with the highest overlap across MCCOP, the genetic algorithm, and stochastic hill climbing. Figure~\ref{fig:properties-ed3} shows the same physicochemical property distributions as Figure~\ref{fig:properties} in the main text, restricted to this subset. The trends observed in the main text are preserved: MCCOP-generated counterfactuals remain within the distribution of the original test set for pLDDT, GRAVY, instability index, and radius of gyration, whereas the discrete baselines show broader deviations, particularly in instability index and GRAVY.

\section{Hyperparameter Sensitivity}
\label{app:hyperparameters}

Table~\ref{tab:hyperparams} lists the primary hyperparameters for MCCOP and the values used. We use the same set of hyperparameters across all three datasets.

\begin{table}[h]
\caption{MCCOP hyperparameters and their values across benchmarks.}
\label{tab:hyperparams}
\centering
\begin{tabular}{llccc}
\toprule
\textbf{Symbol} & \textbf{Description} & \textbf{Fluorescence} & \textbf{Stability} & \textbf{Activity} \\
\midrule
$k$ & Top-$k$ masked positions & 5 & 5 & 5 \\
$\lambda_{\text{dist}}$ & $L_2$ distance weight & 0.1 & 0.1 & 0.1 \\
$m$ & Margin in $\mathcal{L}_{\text{margin}}$ & 2.2 & 2.2 & 2.2 \\
$\alpha$ & Projection strength & 0.3 & 0.3 & 0.3 \\
$t_{\text{diff}}$ & Diffusion noise level & 100 & 100 & 100 \\
$\eta$ & Learning rate & $0.5$ & $0.5$ & $0.5$ \\
$T_{\max}$ & Max optimization steps & 50 & 50 & 50 \\
\bottomrule
\end{tabular}
\end{table}

We perform an ablation over all combinations of the previously mentioned smoother components (including no smoothing) as well as the manifold projection step and the masking value $k$. No smoothing, no masking and no manifold projection corresponds exactly to the gradient descent baseline. Lower $k$ values proved to be too restrictive and caused low success rates while higher ones provided only a marginal increase in success rate. No smoothing significantly increases adversarial rates, while no manifold projection significantly reduces pLDDT scores of generated counterfactuals.

\section{Baselines Implementation Details}
\label{app:baslines}
We compare our method against three baseline counterfactual explanation strategies, each operating over protein sequences and their corresponding embeddings. All baselines share a common interface and are evaluated using the same predictor and confidence threshold ($\tau = 0.95$ by default). Below we describe each baseline along with its hyperparameters.

\subsection{Gradient Descent}
\label{app:baseline-gd}

This baseline performs standard gradient descent directly in the continuous embedding space. Given an input embedding $\mathbf{x}$, a differentiable copy $\mathbf{x}'$ is optimized to maximize the predictor's probability of the target (flipped) class via binary cross-entropy loss. The Adam optimizer is used to update $\mathbf{x}'$ over a fixed number of steps. At each step, the candidate counterfactual with the highest confidence toward the target class is retained.

\begin{table}[h]
\centering
\caption{Hyperparameters for the gradient descent baseline.}
\label{tab:hp-gradient-ascent}
\begin{tabular}{lcc}
\toprule
\textbf{Hyperparameter} & \textbf{Symbol} & \textbf{Value} \\
\midrule
Learning rate          & $\eta$        & $1 \times 10^{-2}$ \\
Gradient steps         & $T$           & $50$ \\
Confidence threshold   & $\tau$        & $0.95$ \\
Optimizer              & --            & Adam \\
Loss function          & $\mathcal{L}$ & BCEWithLogitsLoss \\
\bottomrule
\end{tabular}
\end{table}

\noindent Notably, this baseline operates entirely in embedding space and does not enforce any manifold constraints or discrete sequence validity, making it a purely continuous relaxation approach.

\subsection{Random Mutation}
\label{app:baseline-rm}

This baseline performs a stochastic hill-climbing search in discrete sequence space. At each step, a single random point mutation is applied to each unsolved sequence: a uniformly random position is selected and replaced with a uniformly random amino acid from the standard 20-letter alphabet. The mutated sequence is re-encoded into embedding space using a lightweight encoder, and the predictor evaluates the new embedding. If the target-class confidence improves, the mutation is accepted; otherwise, the sequence reverts to the previous best. The process repeats for a fixed number of steps.

\begin{table}[h]
\centering
\caption{Hyperparameters for the Random Mutation baseline.}
\label{tab:hp-random-mutation}
\begin{tabular}{lcc}
\toprule
\textbf{Hyperparameter} & \textbf{Symbol} & \textbf{Value} \\
\midrule
Maximum steps          & $T$      & $50$ \\
Confidence threshold   & $\tau$   & $0.95$ \\
Amino acid alphabet    & $\mathcal{A}$ & Standard 20 \\
Mutations per step     & --       & $1$ \\
\bottomrule
\end{tabular}
\end{table}

\subsection{Genetic Algorithm}
\label{app:baseline-gea}

This baseline employs a population-based evolutionary strategy operating in discrete sequence space. For each input sequence, an initial population is constructed by applying random mutations to the original. At each generation, individuals are evaluated by encoding them into embedding space and computing a fitness score defined as the predictor's target-class confidence, optionally penalized by the Hamming distance to the original sequence:
\begin{equation}
    f(\mathbf{s}) = \text{conf}(\mathbf{s}) - \lambda \cdot d_H(\mathbf{s}, \mathbf{s}_{\text{orig}}),
\end{equation}
where $\text{conf}(\mathbf{s})$ is the predictor's confidence on the target class for sequence $\mathbf{s}$, $d_H$ denotes the Hamming distance, and $\lambda$ is the edit distance penalty. Selection uses tournament selection with tournament size 3. The top 20\% of the population is preserved as elites. Offspring are generated via single-point crossover and random point mutation (1--2 mutations per offspring). Evolution proceeds for a fixed number of generations or until all samples exceed the confidence threshold.

\begin{table}[h]
\centering
\caption{Hyperparameters for the Genetic Algorithm baseline.}
\label{tab:hp-genetic-algorithm}
\begin{tabular}{lcc}
\toprule
\textbf{Hyperparameter} & \textbf{Symbol} & \textbf{Value} \\
\midrule
Population size         & $N$          & $40$ \\
Generations             & $G$          & $30$ \\
Crossover rate          & $p_c$        & $0.5$ \\
Edit distance penalty   & $\lambda$    & $0.02$ \\
Confidence threshold    & $\tau$       & $0.95$ \\
Elite fraction          & --           & $20\%$ \\
Tournament size         & $k$          & $3$ \\
Mutations per offspring & --           & $1$--$2$ \\
Maximum batch size      & --           & $8$ \\
\bottomrule
\end{tabular}
\end{table}

\section{Additional Structural Visualizations}
\label{app:structures}
We provide two additional structure visualizations for the stability dataset as we could not verify hydrophobic core packing by investigating mutation frequencies per residue.

\begin{figure}[h]
\centering
\includegraphics[width=0.9\linewidth]{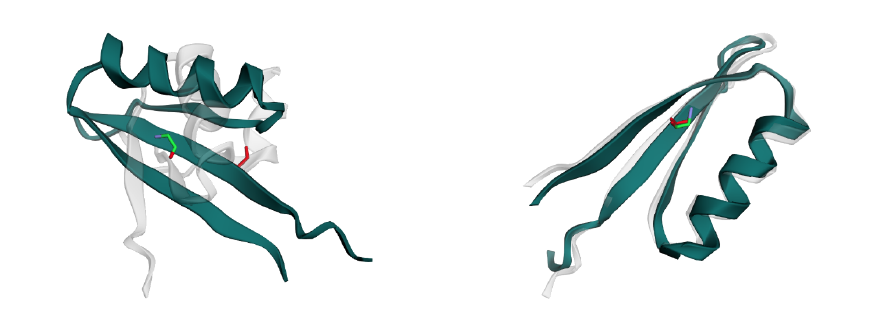}
\caption{Structural alignment of original (gray) and counterfactual (cyan) stability variants across three topologies. Core-facing mutated residues shown as sticks.}
\label{fig:stability-structures}
\end{figure}

\section{Dataset Statistics and Preprocessing}
\label{app:datasets}

\begin{table}[h]
\caption{Dataset statistics after preprocessing.}
\label{tab:dataset-stats}
\centering
\begin{tabular}{lccccc}
\toprule
\textbf{Dataset} & \textbf{Sequences} & \textbf{Positive} & \textbf{Negative} & \textbf{Avg.\ Length} & \textbf{Binarization} \\
\midrule
Fluorescence & 54,025 & 30,697 & 23,328 & 236.96 & Bimodal split \\
Stability & 45,901 & 22,694 & 23,207 & 45.06 & Remove middle 33\% \\
Activity & 60,692 & 30,293 & 30,399 & 102 & Remove middle 33\% \\
\bottomrule
\end{tabular}
\end{table}

For the fluorescence dataset, we exploit the natural bimodality of the log-fluorescence distribution and determine the optimal threshold using Otsu's method. For the stability and activity datasets, removing the middle tercile ensures a clear margin between classes, reducing label noise near the decision boundary. All embeddings are computed using the CHEAP encoder with ESMFold as the backbone, producing representations of dimension $D = 1024$ with no compression along the length dimension.

\end{document}